\documentclass[10pt,twocolumn,letterpaper]{article}

\usepackage{iccv}
\usepackage{times}
\usepackage{epsfig}
\usepackage{graphicx}
\usepackage{amsmath}
\usepackage{amssymb}
\usepackage[dvipsnames]{xcolor}
\usepackage{multirow}
\usepackage{rotating}
\usepackage{xcolor}
\usepackage{soul}

\usepackage[pagebackref=true,breaklinks=true,letterpaper=true,colorlinks,bookmarks=false]{hyperref}

\iccvfinalcopy % *** Uncomment this line for the final submission

\newcommand{\qsection}[1]{\vspace{5pt} \noindent \textbf{#1:}}

\ificcvfinal\fi
\begin{document}

%%%%%%%%% TITLE
\title{Need for Speed: A Benchmark for Higher Frame Rate Object Tracking }

\author{%
\begin{tabular}[t]{c@{\extracolsep{4em}}c}
   \multicolumn{2}{c}{Hamed Kiani Galoogahi$^{1}$\thanks{Kiani \& Fagg are joint first authors.\newline Email: \href{mailto:hamedkg@gmail.com}{hamedkg@gmail.com} \& \href{mailto:ashton@fagg.id.au}{ashton@fagg.id.au}}, Ashton Fagg$^{2,1*}$, Chen Huang$^{1}$, Deva Ramanan$^{1}$ and Simon Lucey$^{1,2}$}\\
   \rule{0pt}{3ex}
   $^1$Robotics Institute & $^2$ SAIVT Lab\\ 
   Carnegie Mellon University & Queensland University of Technology\\
\end{tabular}
}

\newcommand{\myul}[2][black]{\setulcolor{#1}\ul{#2}\setulcolor{black}}

\maketitle

\begin{abstract}
  In this paper, we propose the first higher frame rate video dataset (called Need for Speed - NfS) and benchmark for visual object tracking. The dataset consists of 100 videos (380K frames) captured with now commonly available higher frame rate (240 FPS) cameras from real world scenarios. All frames are annotated with axis aligned bounding boxes and all sequences are manually labelled with nine visual attributes -  such as occlusion, fast motion, background clutter, etc. Our benchmark provides an extensive evaluation of many recent and state-of-the-art trackers on higher frame rate sequences. We ranked each of these trackers according to their tracking accuracy and real-time performance. One of our surprising conclusions is that at higher frame rates, simple trackers such as correlation filters outperform complex methods based on deep networks. This suggests that for practical applications (such as in robotics or embedded vision), one needs to carefully tradeoff bandwidth constraints associated with higher frame rate acquisition, computational costs of real-time analysis, and the required application accuracy. Our dataset and benchmark allows for the first time (to our knowledge) systematic exploration of such issues, and will be made available to allow for further research in this space.
  
\end{abstract}

%%%%%%%%% BODY TEXT
\section{Introduction}

\begin{figure}
\begin{center}
\begin{tabular}{c}
   \includegraphics[width=.47\textwidth]{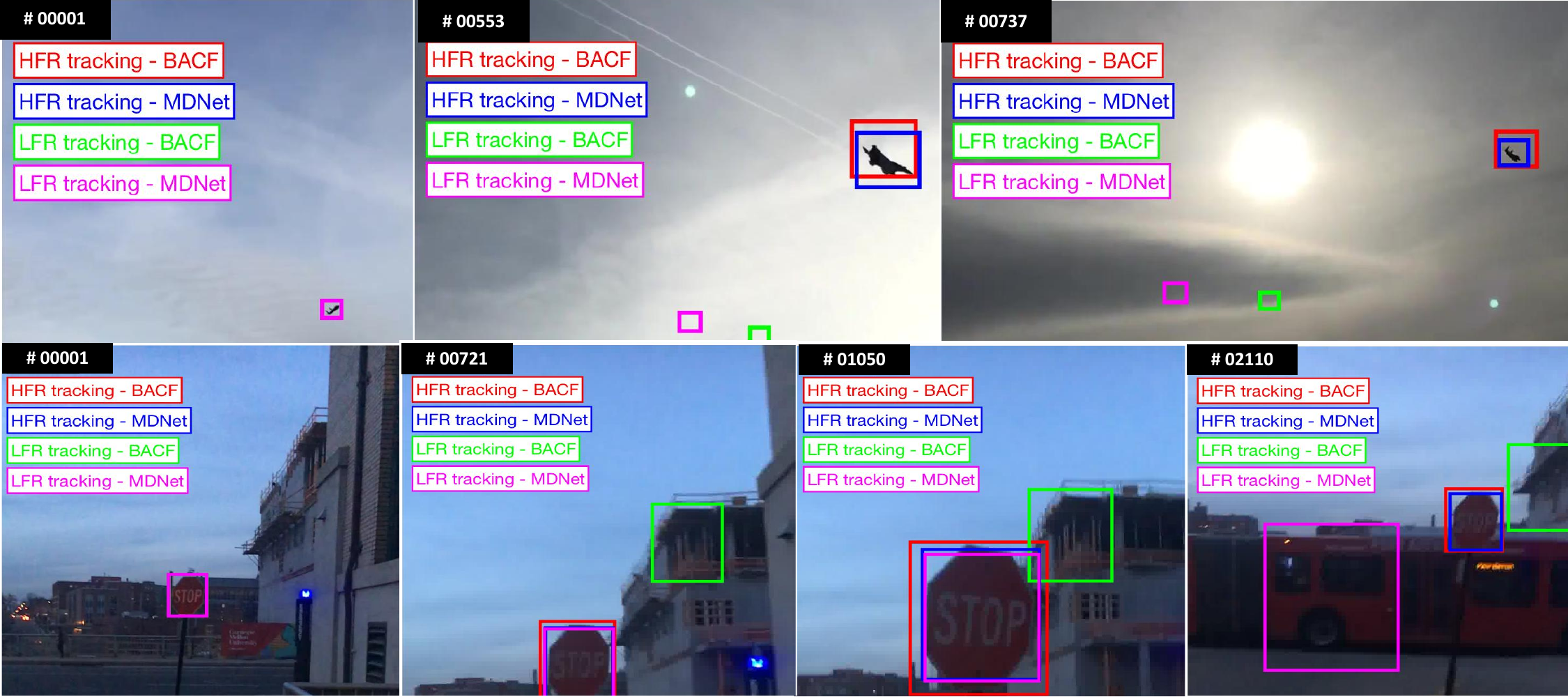}   \\ 
     \includegraphics[width=.47\textwidth]{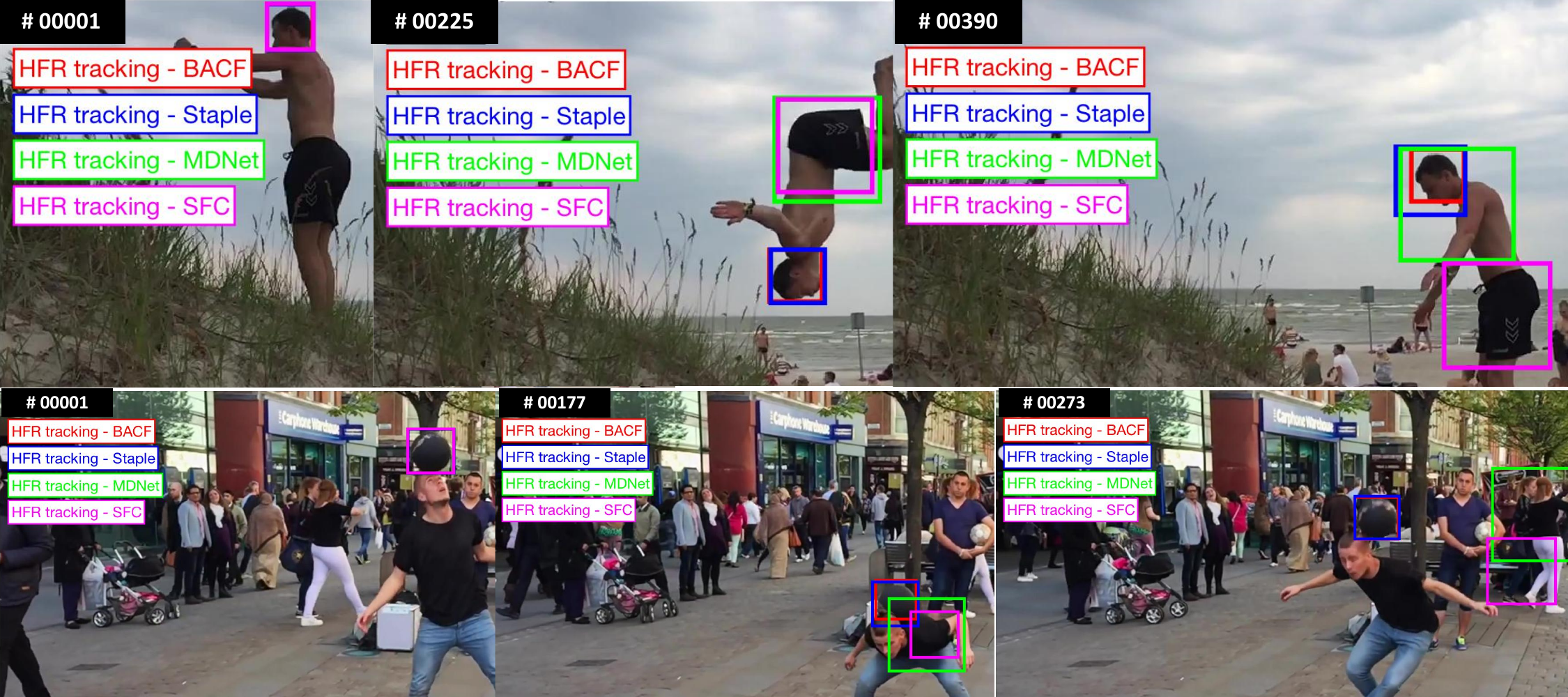}
\end{tabular}
\end{center}
            \caption{The effect of tracking higher frame rate videos. Top rows illustrate the robustness of tracking higher frame rate videos (240 FPS) versus lower frame rate videos (30 FPS) for a Correlation Filter (BACF with HOG) and a deep tracker (MDNet). Bottom rows show if higher frame rate videos are available, cheap CF trackers (Staple and BACF) can outperform complicated deep trackers (SFC and MDNet) on challenging situations such as fast motion, rotation, illumination and cluttered background. Predicted bounding boxes of these methods are shown by different colors. HFR and LFR refer to Higher and Lower Frame Rate videos.}
\label{fig:intro}
\end{figure}

Visual object tracking is a fundamental task in computer vision which has implications for a bevy of applications: surveillance, vehicle autonomy, video analysis, etc. The vision community has shown an increasing degree of interest in the problem - with recent methods becoming increasingly sophisticated and accurate~\cite{danelljan2015learning, henriques2015high, danelljan2014adaptive, bertinetto2015staple, bertinetto2016fully}. However, most of these algorithms - and the datasets they have been evaluated upon~\cite{wu2015object, wu2013online, liang2015encoding} - have been aimed at the canonical approximate frame rate of 30 Frames Per Second (FPS). Consumer devices with cameras such as smart phones, tablets, drones, and robots are increasingly coming with higher frame rate cameras (240 FPS now being standard on many smart phones, tablets, drones, etc.). The visual object tracking community is yet to adapt to the changing landscape of what ``real-time'' means and how faster frame rates effect the choice of tracking algorithm one should employ. 

In recent years, significant attention has been paid to Correlation Filter (CF) based methods ~\cite{bolme2010visual, henriques2015high, danelljan2015learning, kiani2013multi, kiani2015correlation, bertinetto2015staple} for visual tracking. The appeal of correlation filters is their efficiency - a discriminative tracker can be learned online from a single frame and adapted after each subsequent frame. This online adaptation process allows for impressive tracking performance from a relatively low capacity learner. Further, CFs take advantage of intrinsic computational efficiencies afforded from operating in the Fourier domain~\cite{Kumar2005}. Some CF methods (such as~\cite{henriques2015high,bolme2010visual}) are able to operate at hundreds of frames per second on embedded devices.

More recently, however, the visual tracking community has started to focus upon improving reliability and robustness through advances in deep learning~\cite{bertinetto2016fully, tao2016siamese, wang2015visual}. While such methods have been shown to work well, their use comes at a cost. First, extracting discriminative features from CNNs or applying deep tracking frameworks is computationally expensive. Some deep methods operate at only a fraction of a frame per second, or require a high-end GPU to achieve real time performance. Second, training deep trackers can sometimes require a very large amount of data, as the learners are high capacity and require a significant amount of expense to train.

It is well understood that the central artifact that effects a visual tracker's performance is tolerance to appearance variation from frame to frame. Deep tracking methods have shown a remarkable aptitude for providing such tolerance - with the unfortunate drawback of having a considerable computational footprint~\cite{tao2016siamese, wang2015visual}. In this paper we want to explore an alternate thesis. Specifically, we want to explore that if we actually increase the frame rate - thus reducing the amount of appearance variation per frame - could we get away with substantially simpler tracking algorithms (from a computational perspective)? Further, could these computational savings be dramatic enough to cover the obvious additional cost of having to process a significant number of more image frames per second. Due to the widespread availability of higher frame rate (e.g. 240 FPS) cameras on many consumer devices we believe the time is ripe to explore such a question.

Inspired by the recent work of Handa et al.~\cite{handa2012real} in visual odometry, we believe that increasing capture frame rate allows for a significant boost in tracking performance without the need for deep features or complex trackers. We do not dismiss deep trackers or deep features, however we show that under some circumstances they are not necessary. By trading tracker capacity for higher frame rates - as is possible in many consumer devices - we believe that more favourable runtime performance can be obtained, particularly on devices with resource constraints, while still obtaining competitive tracking accuracy.

\qsection{Contributions} In this paper, we present the Need for Speed (NfS) dataset and benchmark, the first benchmark (to our knowledge) for higher frame rate general object tracking using consumer devices. We use our dataset to evaluate numerous state of the art tracking methods (both CFs and deep learning based methods). An exciting outcome of our work was the unexpected result that if a sufficiently higher frame rate can be attained CFs with cheap hand-crafted features (\eg HOG~\cite{hog}) can outperform state of the art deep trackers in terms of accuracy and computational efficiency.  

%%%%%%%%%%%%%%%%%%%%%%%%%%%%%%%%%%%%%%%%%%%%%%%%%%%%%%%%%%%%%%%%%%%%%%%%%%%%%%%%%%%%%%%%%%%%%%%%%%%%%%
%%%%%%%%%%%%%%%%%%%%%%%%%%%%%               Related Work %%%%%%%%%%%%%%%%%%%%%%%%%%%%%        
%%%%%%%%%%%%%%%%%%%%%%%%%%%%%%%%%%%%%%%%%%%%%%%%%%%%%%%%%%%%%%%%%%%%%%%%%%%%%%%%%%%%%%%%%%%%%%%%%%%%%%

\section{Related Work}

\subsection{Tracking Datasets}
Standard tracking datasets such as OTB100~\cite{wu2015object}, VOT14~\cite{kristan2014visual} and ALOV300~\cite{smeulders2014visual} have been widely used to evaluate current tracking methods in the literature. These datasets display annotated generic objects in real-world scenarios captured by low frame rate cameras (\ie 24-30 FPS). Existing tracking datasets are briefly described as below.

\qsection{OTB50 and OTB100} OTB50~\cite{wu2013online} and OTB100~\cite{wu2015object} belong to the Object Tracking Benchmark (OTB) with 50 and 100 sequences, respectively. OTB50 is a subset of OTB100, and both datasets are annotated with bounding boxes as well as 11 different attributes such as illumination variation, scale variation and occlusion and deformation.

\qsection{Temple-Color 128 (TC128)} This dataset consists of 128 videos which was specifically designed for the evaluation of color-enhanced trackers. Similar to OTBs, TC128 provides per frame bounding boxes and 11 per video attributes~\cite{liang2015encoding}.

\qsection{VOT14 and VOT15} VOT14 ~\cite{kristan2014visual} and VOT15 ~\cite{kristan2015visual} consist of 25 and 30 challenging videos, respectively, which are mainly borrowed from OTB100. All videos are labelled with rotated bounding boxes rather than upright ones. Both datasets come with per frame attribute annotation. 

\qsection{ALOV300} This dataset ~\cite{alov} contains 314 sequences mainly borrowed from the OTBs, VOT challenges and TC128. Videos are labeled with 14 visual attributes such as low contrast, long duration, confusion, zooming camera, motion smoothness, moving camera and transparency.

\qsection{UAV123} This dataset~\cite{mueller2016benchmark} is recently created for Unmanned Aerial Vehicle (UAV) target tracking. There are 128 videos in this dataset, 115 videos captured by UAV cameras and 8 sequences rendered by a UAV simulator, which are all annotated with bounding boxes and 12 attributes.

Table~\ref{table:stats} compares the NfS dataset with these datasets, showing that NfS is the only dataset with higher frame rate videos captured at 240 FPS. Moreover, in terms of number of frames, NfS is the largest dataset with 380K frames which is more than two times bigger than ALOV300. 

\begin{table*}
\centering
\caption{Comparing NfS with other object tracking datasets. }
\label{table:stats}
\begin{tabular}{lcccccccccc}
\hline
& UAV123 & OTB50 & OTB100 & TC128 & VOT14 & VOT15 & ALOV300 & NfS\\
& ~\cite{mueller2016benchmark} & ~\cite{wu2013online} & ~\cite{wu2015object} & ~\cite{liang2015encoding} & ~\cite{kristan2014visual} &~\cite{kristan2015visual} & ~\cite{alov} &\\
\hline
Capture frame rate   & 30      & 30    & 30     & 30    & 30    & 30    & 30      & 240    \\  
\# Videos    & 123       & 51    & 100    & 129   & 25    & 60    & 314     & 100    \\  
Min frames   & 109        & 71    & 71     & 71    & 171   & 48    & 19      & 169    \\ 
Mean frames  & 915      & 578   & 590    & 429   & 416   & 365   & 483     & 3830   \\  
Max frames   & 3085      & 3872  & 3872   & 3872  & 1217  & 1507  & 5975    & 20665  \\  
Total frames & 112578    & 29491 & 59040  & 55346 & 10389 & 21871 & 151657  & 383000 \\ \hline
\end{tabular}
\end{table*}

\subsection{Tracking Methods}
 
Recent trackers can be generally divided into two categories, including correlation filter (CF) trackers~\cite{bertinetto2015staple, henriques2015high, danelljan2015convolutional,ma2015hierarchical, danelljan2016beyond} and deep trackers~\cite{nam2015learning,bertinetto2016fully,wang2016stct,tao2016siamese}. We briefly review each of these two categories as following.

\qsection{Correlation Filter Trackers} The interest in employing CFs for visual tracking was ignited by the seminal MOSSE filter~\cite{bolme2010visual} with an impressive speed of $\sim$700 FPS, and the capability of online adaptation. Thereafter, several works~\cite{henriques2015high,bertinetto2015staple,danelljan2014accurate,danelljan2014adaptive,ma2015long} were built upon the MOSSE showing notable improvement by learning CF trackers from more discriminative multi-channel features (\eg HOG~\cite{hog}) rather than pixel values. 
KCF ~\cite{henriques2015high} significantly improved MOSSE's accuracy by real-time learning of kernelized CF trackers on HOG features. Trackers such as Staple~\cite{bertinetto2015staple}, LCT~\cite{ma2015long} and SAMF~\cite{danelljan2014adaptive} were developed to improve KCF's robustness to object deformation and scale change. Kiani \textit{et al.}~\cite{kiani2015correlation} showed that learning such trackers in the frequency domain is highly affected by boundary effects, leading to suboptimal performance~\cite{danelljan2015learning}. The CF with Limited Boundaries (CFLB) ~\cite{kiani2015correlation}, Spatially Regularized CF (SRDCF) ~\cite{danelljan2015learning} and the Background-Aware CF (BACF)~\cite{bacf} have proposed solutions to mitigate these boundary effects in the Fourier domain, with impressive results.

Recently, learning CF trackers from deep Convolutional Neural Networks (CNNs) features~\cite{simonyan2014very,krizhevsky2012imagenet} has offered superior results on several standard tracking datasets~\cite{danelljan2016beyond,danelljan2015convolutional,ma2015hierarchical}. The central tenet of these approaches (such as HCF~\cite{ma2015hierarchical} and HDT~\cite{qi2016hedged}) is that- even by per frame online adaptation- hand-crafted features such as HOG are not discriminative enough to capture the visual difference between consecutive frames in low frame rate videos. Despite their notable improvement, the major drawback of such CF trackers is their intractable complexity ($\sim$0.2 FPS on CPUs) mainly needed for extracting deep features and computing Fourier transforms on hundreds of feature channels. 

\qsection{Deep Trackers} Recent deep learning based trackers ~\cite{NIPS2013_5192, nam2015learning, wang2015visual} represent a new paradigm in tracking. Instead of hand-crafted features, a deep network trained for a non-tracking task (such as object recognition~\cite{krizhevsky2012imagenet}) is updated with video data for generic object tracking. Unlike the direct combination of deep features with the traditional shallow methods~\eg~CFs~\cite{danelljan2016beyond,ma2015hierarchical,danelljan2015convolutional,qi2016hedged}, the updated deep trackers aim to learn from scratch the target-specific features for each new video. For example, the MDNet~\cite{nam2015learning} learns generic features on a large set of videos and updates the so called domain-specific layers for unseen ones. The more related training set and unified training and testing approaches make MDNet win the first place in the VOT15 Challenge. Wang~\etal~\cite{wang2015visual} proposed to use fully convolutional networks (FCNT) with feature map selection mechanism to improve performance. However, such methods are computationally very expensive (even with a high end GPU) due to the fine-tuning step required to adapt the network from a large number of example frames. There are two high-speed deep trackers GOTURN~\cite{held2016} and SFC~\cite{bertinetto2016fully} that are able to run at 100 FPS and 75 FPS respectively on GPUs. Both of these methods train a Siamese network offline to predict motion between two frames (either using deep regression or a similarity comparison). At test time, the network is evaluated without any fine-tuning. Thus, these trackers are significantly less expensive because the only computational cost is the fixed feed-forward process. For these trackers, however, we remark that there are two major drawbacks. First, their simplicity and fixed-model nature can lead to high speed, but also lose the ability to update the appearance model online which is often critical to account for drastic appearance changes. Second, on modern CPUs, their speed becomes no more than 3 FPS, which is too slow for practical use on devices with limited computational resources.

%%%%%%%%%%%%%%%%%%%%%%%%%%%%%%%%%%%%%%%%%%%%%%%%%%%%%%%%%%%%%%%%%%%%%%%%%%%%%%%%%%%%%%%%%%%%%%%%%%%%%%
%%%%%%%%%%%%%%%%%%%%%%%%%%%%%               HFR Benchmark  %%%%%%%%%%%%%%%%%%%%%%%%%%%%%        
%%%%%%%%%%%%%%%%%%%%%%%%%%%%%%%%%%%%%%%%%%%%%%%%%%%%%%%%%%%%%%%%%%%%%%%%%%%%%%%%%%%%%%%%%%%%%%%%%%%%%%

\section{NfS Dataset} \label{sec:nfs_dataset}

The NfS datset consists of 100 higher frame rate videos captured at 240 FPS. We captured 75 videos using the iPhone 6 (and above) and the iPad Pro which are capable of capturing 240 frames per second. We also included 25 sequences from YouTube, which were captured at 240 FPS from a variety of different devices. All 75 captured videos come with corresponding IMU and Gyroscope raw data gathered during the video capturing process. Although we make no use of such data in this paper, we will make the data publicly available for potential applications.

The tracking targets include (but not limited to) vehicle (bicycle, motorcycle, car), person, face, animal (fish, bird, mammal, insect), aircraft (airplane, helicopter, drone), boat, and generic objects (\eg sport ball, cup, bag, etc.). Each frame in NfS is annotated with an axis aligned bounding box using the VATIC toolbox ~\cite{vatic}. Moreover, all videos are labeled with nine visual attributes, including occlusion, illumination variation (IV), scale variation (SV), object deformation (DEF), fast motion (FM), viewpoint change 
(VC), out of view (OV), background clutter (BC) and low resolution (LR). The distribution of these attributes for NfS is presented in Table ~\ref{table_attr_dist}. Example frames of the NfS dataset and detailed description of each attribute are provided in the supplementary material.

%%%%%%%%%%%%%%%%%%%%%%%%%%%%%%%%%%%%%%%%%%%%%%%%%%%%%%%%%%%%%%%%%%%%%%%%%%%%%%%%%%%%%%%%%%%%%%%%%%%%%%
%%%%%%%%%%%%%%%%%%%%%%%%%%%%%               Evaluation %%%%%%%%%%%%%%%%%%%%%%%%%%%%%        
%%%%%%%%%%%%%%%%%%%%%%%%%%%%%%%%%%%%%%%%%%%%%%%%%%%%%%%%%%%%%%%%%%%%%%%%%%%%%%%%%%%%%%%%%%%%%%%%%%%%%%
\section{Evaluation}

\qsection{Evaluated Algorithms} We evaluated 15 recent trackers on the NfS dataset. We generally categorised these trackers based on their learning strategy and utilized feature in three classes including CF trackers with hand-crafted features (BACF~\cite{bacf}, SRDCF~\cite{danelljan2015learning}, Staple~\cite{bertinetto2015staple}, DSST~\cite{danelljan2014accurate}, KCF~\cite{henriques2015high}, LCT~\cite{ma2015long}, SAMF~\cite{li2014scale} and CFLB~\cite{kiani2015correlation}), CF trackers with deep features (HCF~\cite{ma2015hierarchical} and HDT~\cite{qi2016hedged}) and deep trackers (MDNet~\cite{nam2015learning}, SiameseFc~\cite{bertinetto2016fully}, FCNT~\cite{wang2015visual}, GOTURN~\cite{held2016}). We also included MEEM~\cite{zhang2014meem} in the evaluation as the state of the art SVM-based tracker with hand-crafted feature. All these trackers are detailed in the supplementary material in terms of learning strategy and feature representation.

\begin{table}
\centering
%\caption{Attribute distribution}
\caption{Distribution of visual attributes within the NfS dataset, showing the number of coincident attributes across all videos. Please refer to Section \hyperref[sec:nfs_dataset]{3} for more details.} 
\label{table_attr_dist}
\begin{tabular}{@{\hskip .67em}l@{\hskip .67em}c@{\hskip .67em}c@{\hskip .67em}c@{\hskip .67em}c@{\hskip .67em}c@{\hskip .67em}c@{\hskip .67em}c@{\hskip .67em}c@{\hskip .67em}c}
\hline 
~ & IV & SV  & OCC & DEF &  FM & VC & OV & BC & LR \\
\hline 
IV & \textbf{45}  &  39 &   23    &12  &   33 &   17  &   9  &  16    & 8 \\
SV & 39  &  \textbf{83}  &  41  &  36      & 57  &  41   & 21  &  28  &   8 \\
OCC & 23 &   41  &  \textbf{51}  &  21    &  31  &  23  &  10  &  19   &  8 \\
DEF & 12 &   36  &  21  &  \textbf{38}     &  19  &  30   &  4  &  13   &  1 \\
%AFB & 10  &  14  &   9  &   6  &  \textbf{20}  &  12  &   7  &   5  &  10 &    5 \\
FM & 33  &  57 &   31  &  19  &   \textbf{70} &   31  &  20  &  24  &   4 \\
VC&17  &  41 &   23 &   30      &31 &   \textbf{46}  &   9  &  16 &    0 \\
OV&9  &  21  &  10   &  4      &20   &  9  &  \textbf{22}  &  10   &  4 \\
BC&16   & 28  &  19  &  13     & 24  &  16  &  10 &   \textbf{36}   &  6 \\
LR& 8   &  8   &  8   &  1    & 4   &  0   &  4  &   6  &  \textbf{10}\\
\hline
\end{tabular}
\end{table}

\qsection{Evaluation Methodology} We use the success metric to evaluate all the trackers~\cite{wu2013online}. Success measures the intersection over union (IoU) of predicted and ground truth bounding boxes. The success plot shows the percentage of bounding boxes whose IoU is larger than a given threshold. We use the Area Under the Curve (AUC) of success plots to rank the trackers. We also compare all the trackers by their success rate at the conventional thresholds of 0.50 (IoU $>$ 0.50)~\cite{wu2013online}. Moreover, we report the relative accuracy improvement which is computed as $\frac{\text{improved accuracy}}{\text{accuracy of lower frame rate tracking}}$, where the improved accuracy is the difference between accuracy (success rate at IoU $>$ 0.50) of higher frame rate and lower frame rate tracking.

\begin{figure}
\begin{center}
    \begin{tabular}{c}
    \includegraphics[width=.4\textwidth]{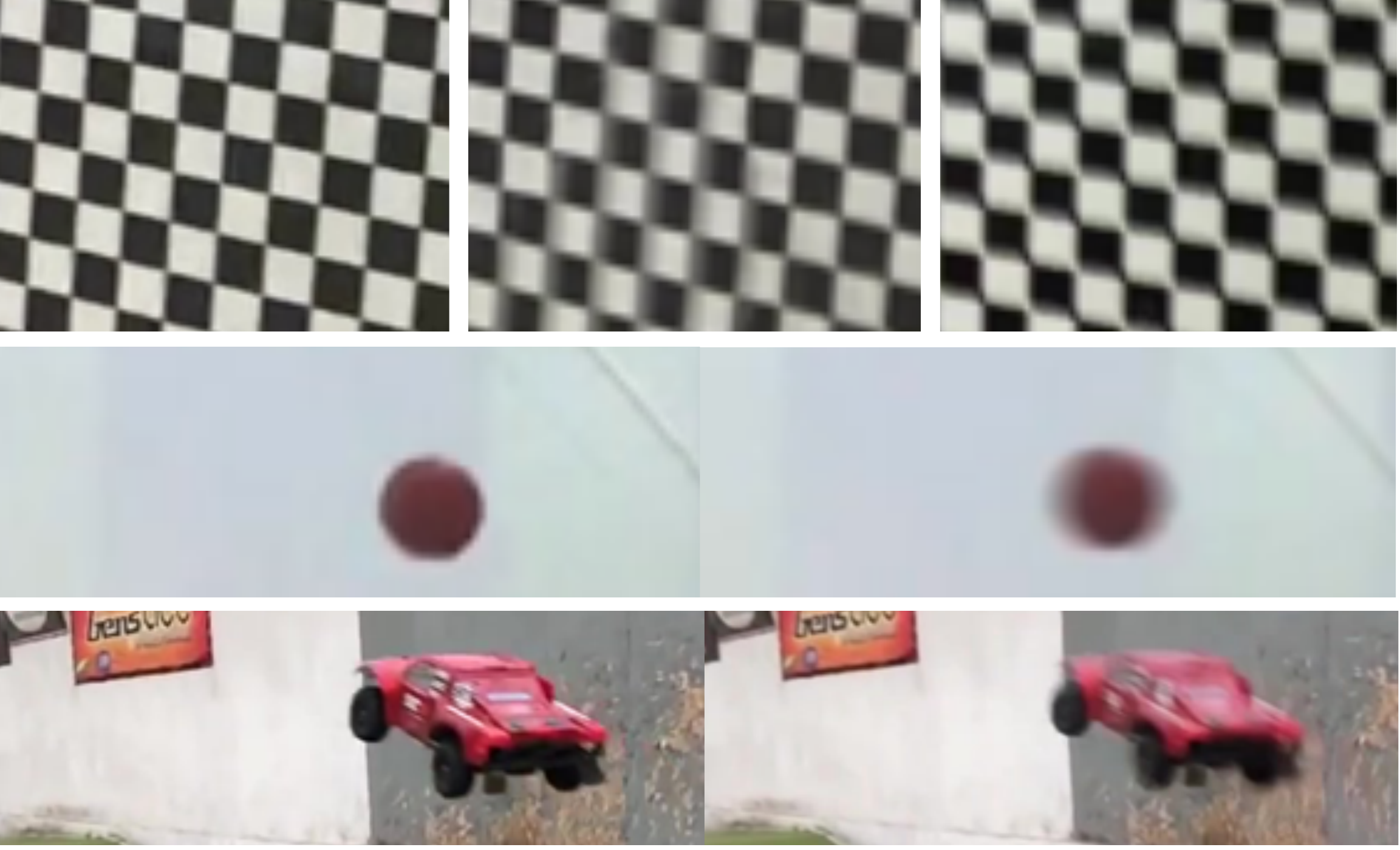} 
    \end{tabular}
\end{center}
            \caption{Top) a frame captured by a high frame rate camera (240 FPS), the same frame with synthesized motion blur, and the same frame captured by a low frame rate camera (30 FPS) with real motion blur. Bottom) sampled frames with corresponding synthesized motion blur. Please refer to Tracking Scenarios for more details.}
\label{fig:mb_Samples}
\end{figure}

\qsection{Tracking Scenarios} To measure the effect of capture frame rate on tracking performance, we consider two different tracking scenarios. At the first scenario, we run each tracker over all frames of the higher frame rate videos (240 FPS) in the NfS dataset. The second scenario, on the other hand, involves tracking lower frame rate videos (30 FPS). Since all videos in the NfS dataset are captured by high frame rate cameras, and thus no 30 FPS video is available, we simply create a lower frame rate version of NfS by temporal sampling every 8th frame. In such case, we track the object over each 8th frame instead of all frames. This simply models the large visual difference between two following sampled frames as one may observe in a real lower frame rate video. However, the main issue of temporal sampling is that since the videos are originally captured by higher frame rate cameras with very short exposure time, the motion blur caused by fast moving object/camera is significantly diminished. This leads to excluding the effect of motion blur in lower frame rate tracking. To address this concern and make the evaluation as realistic as possible, we simulate motion blur over the lower frame rate videos created by temporal sampling. We utilize a leading visual effects package (Adobe After Effects) to synthesize motion blur over the sampled frames. To verify the realism of the synthesized motion blur, Fig.~\ref{fig:mb_Samples} demonstrates a real frame (of a checkerboard) captured by a 240 FPS camera, the same frame with synthesized motion blur and a frame with real motion blur captured by a 30 FPS camera with identical extrinsic and intrinsic settings. To capture two sequences with different frame rates, we put two iPhones capture rates of 30 and 240 FPS side-by-side and then capture sequences from the same scene simultaneously. Fig.~\ref{fig:mb_Samples} also shows two frames before and after adding synthesized motion blur.

\begin{table*}
\centering
\caption{Evaluating the effect of updating learning rate of each CF tracker on tracking higher frame rate videos (240 FPS). Accuracy is reported as success rate ($\%$) at IoU $>$ 0.50. Please refer to Section \hyperref[sec:lr]{4.1}
 for more details about the original and updated learning rates.}
\label{table:LR}
\begin{tabular}{lllllllllll}
\hline 
            & BACF & SRDCF & Staple & LCT  & DSST  & SAMF & KCF & CFLB & HCF  & HDT\\ \hline
Original LR & 48.8 & 48.2  & 51.1 & 34.5 & 44.0 & 42.8 & 28.7  & 18.3   & 33.0 & 57.7 \\
Updated LR  & 60.5 & 55.8  & 53.4 & 36.4 & 53.4 & 51.7 & 34.8  & 22.9   & 41.2 & 59.6 \\ \hline
\end{tabular}
\end{table*}

\begin{figure*}
\begin{center}
    \begin{tabular}{c}
    \includegraphics[width=.85\textwidth]{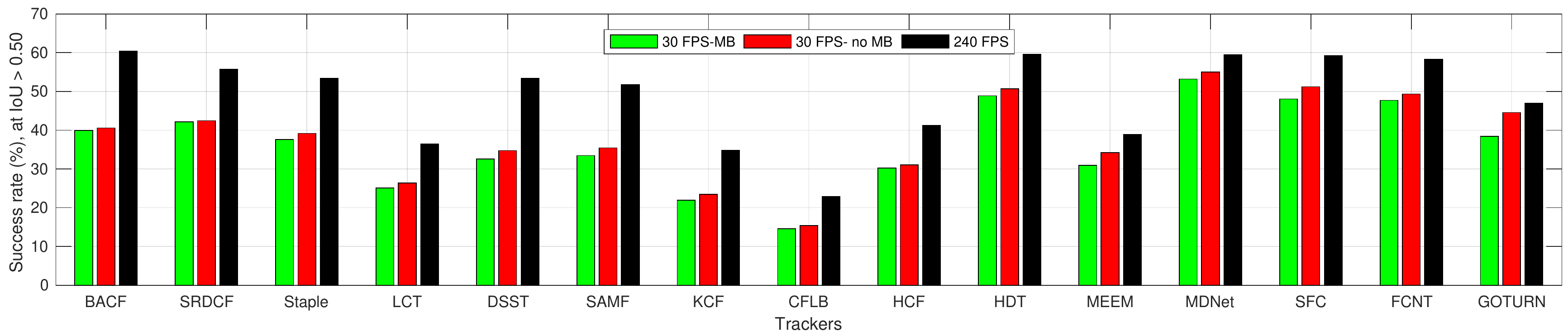} 
    
    \end{tabular}
\end{center}
            \caption{Comparing higher frame rate tracking (240 FPS) versus lower frame rate tracking (30 FPS) for each tracker. For higher frame rate tracking CF trackers employ updated learning rates. The results of lower frame rate tracking are plotted for videos with and without motion blur (30 FPS-MB and 30 FPS- no MB). Results are reported as success rate ($\%$) at IoU $>$ 0.50.}
\label{fig:OP_each_tracker}
\end{figure*}

\subsection{Adjusting Learning Rate of CF Trackers }
\label{sec:lr}
A unique characteristic of CF trackers is their inherent ability to update the tracking model online, when new frames become available. The impact of each frame in learning/updating process is controlled by a constant weight called learning (or adaptation) rate~\cite{bolme2010visual}. Smaller rates increase the impact of older samples, while bigger ones give higher weight to samples from more recent frames. Each CF tracker has its own learning rate which was tuned for robust tracking on low frame rate sequences (30 FPS). To retain the robustness of such methods for higher frame rate videos (240 FPS), we approximately adjust their learning rate to be $LR_{new} = \frac{1}{8} LR_{old}$. Since the number of frames in 240 FPS videos is 8 times more than that in 30 FPS sequences over a fixed period of time, dividing learning rates by 8 can keep the balance between CFs updating capacity and smaller inter-frame variation in 240 FPS videos~\footnote{Proof is provided in the supplementary material.}.

Here, we empirically demonstrate how adjusting the learning rates of CF trackers affects their tracking performance. Table~\ref{table:LR} shows the tracking accuracy (success rate at IoU $>$ 0.50) of 10 recent CF trackers on the NfS 240 FPS sequences, comparing tracking by original learning rates ($LR_{old}$ from their reference papers) versus updated rates $LR_{new}$. The result shows that adjusting the learning rates notably improves the accuracy of all the CF trackers. For some trackers such as BACF, SRDCF, DSST and HCF there is a substantial improvement, while for Staple, LCT and HDT the improvement is much smaller ($\sim 2 \%$). This is most likely because of the complementary parts of these trackers. Staple utilizes color scores per pixel, and LCT uses random fens classifiers as additional detectors/trackers which are independent of their CF modules. Similarly, HDT employs the Hedge algorithm~\cite{chaudhuri2009parameter} as a multi-experts decision maker to merge hundreds of weak CF trackers in a strong tracker. Thus, updating their learning rates offers less improvement compared to those trackers such as BACF and DSST that solely track by a single CF based tracker.  

\begin{figure*}
\begin{center}
    \begin{tabular}{c@{\hskip 0.2em }c@{\hskip 0.2em }c@{\hskip 0.2em }}
    \includegraphics[width=.32\textwidth]{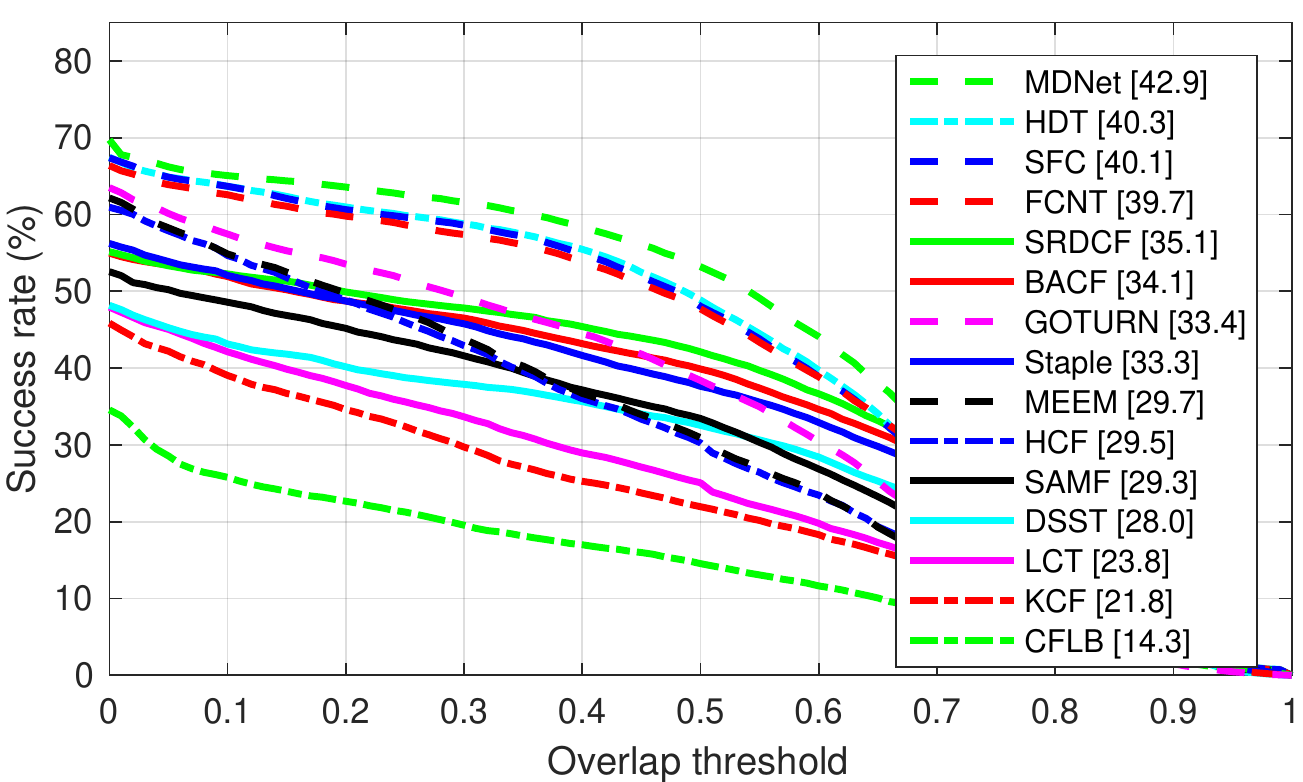} &
    
      \includegraphics[width=.32\textwidth]{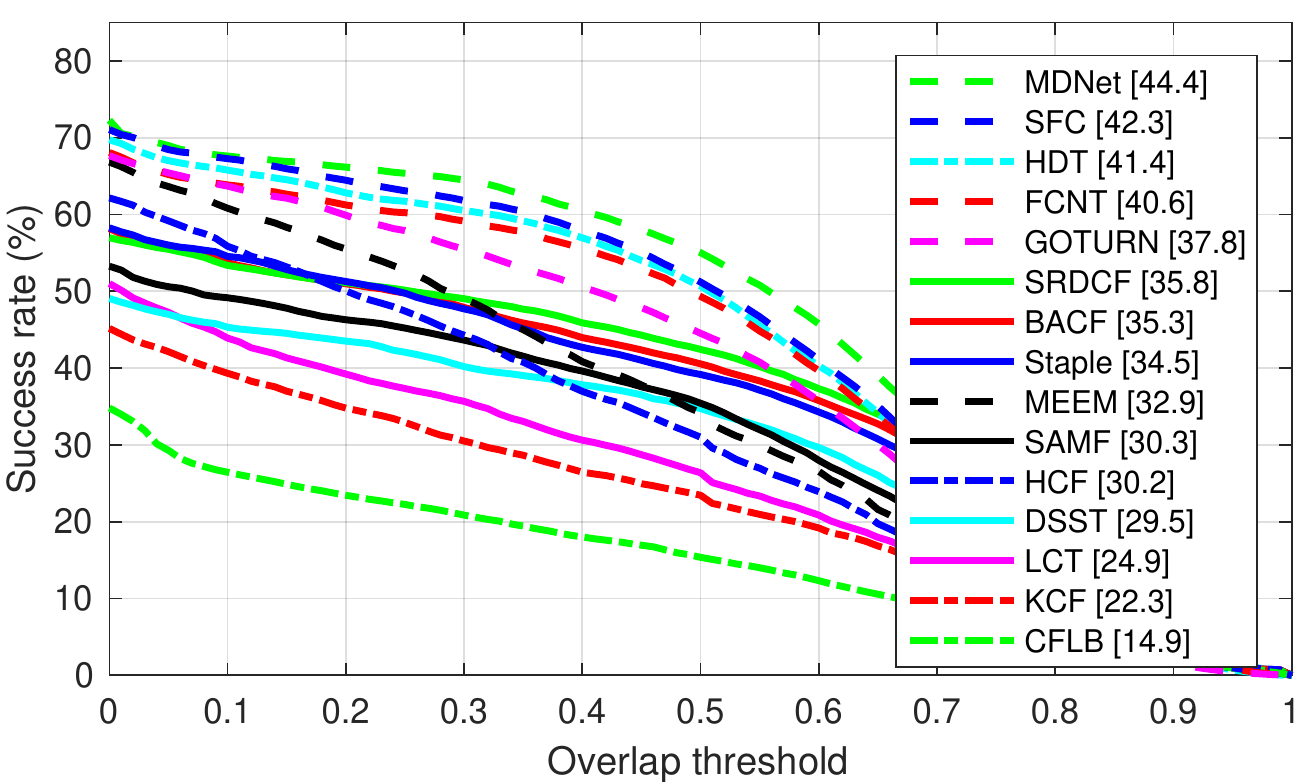} &
   
        \includegraphics[width=.32\textwidth]{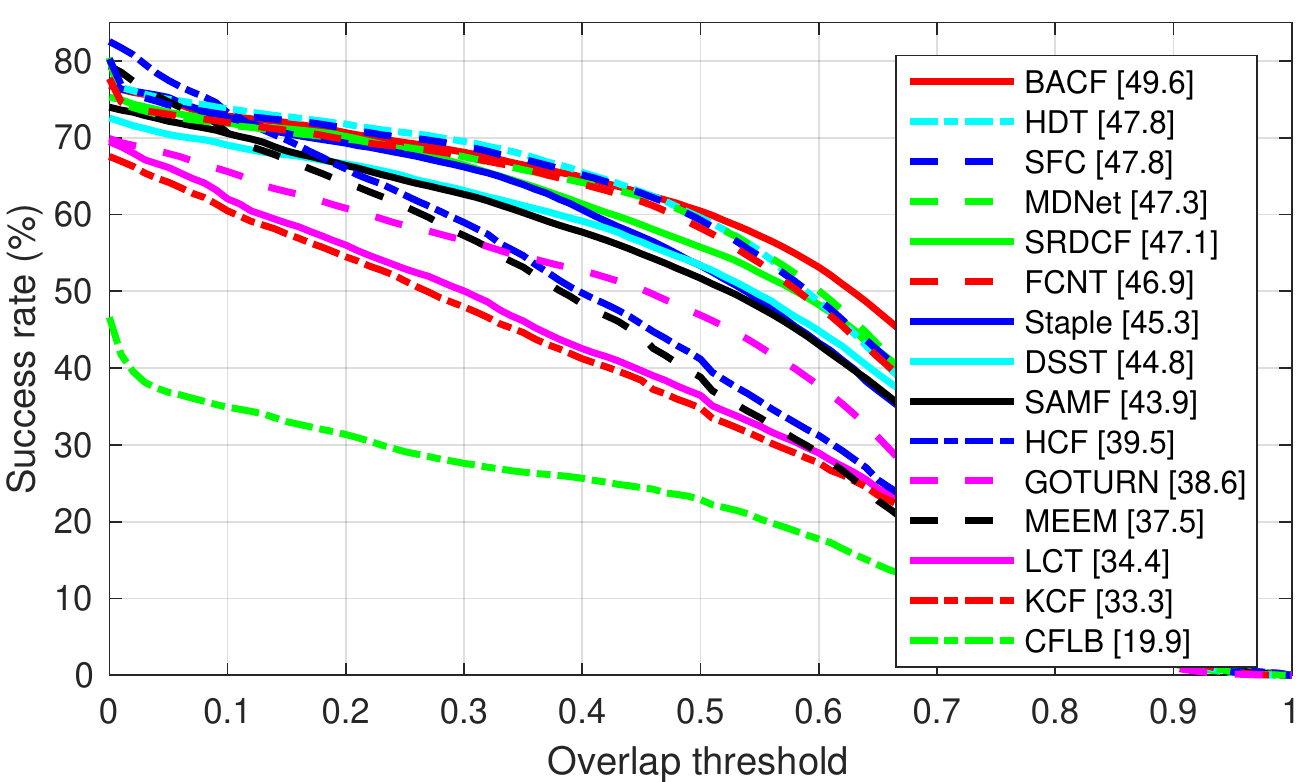} \\
     (a) & (b) & (c)
    
    \end{tabular}
\end{center}
            \caption{Evaluating trackers over three tracking scenarios, (a) lower frame rate tracking with synthesized motion blur, (b) and lower frame rate tracking without motion blur, and (c) higher frame rate tracking. AUCs are reported in brackets.}
\label{fig:OP_sep}
\end{figure*}

\begin{table*}
\centering
\caption{Comparing trackers on three tracking scenarios including higher frame rate tracking (240 FPS), lower frame rate tracking with synthesized motion blur (30 FPS MB) and lower frame rate tracking without motion blur (30 FPS no MB). Results are reported as the AUC of success plots. We also show the speed of each tracker on CPUs and/or GPUs if applicable. The {\color{red}first}, {\color{green}second}, {\color{blue}third}, {\color{cyan}forth} and {\color{Purple}fifth} highest AUCs/speeds are highlighted in color.}
\label{table:all_trackers_AUC}
\resizebox{\linewidth}{!}{
\begin{tabular}{l@{\hskip 0em }c@{\hskip 0.3em }@{}c@{\hskip .3em }@{}c@{\hskip .4em }@{}c@{\hskip .3em }@{}c@{\hskip .3em }@{}c@{\hskip .3em }@{}c@{\hskip .3em }@{}c@{\hskip .3em }@{}c@{\hskip .3em }@{}c@{\hskip .4em }@{}c@{\hskip .4em }@{}c@{\hskip .4em }@{}c@{\hskip .4em }@{}c@{\hskip .4em }@{}c@{\hskip .4em }}
\hline 
                  & BACF & SRDCF & Staple & LCT  & DSST & SAMF & KCF  & CFLB & HCF  & HDT  & MEEM & MDNet & SFC & FCNT & GOTURN\\ \hline
30 FPS- no MB     & 35.2 & {\color{Purple}35.7}  & 34.5& 24.8 & 29.4 & 30.3 & 22.3 & 14.9  & 30.2 & {\color{blue}41.3} & 32.9 & {\color{red}44.4}  & {\color{green}42.3}  & {\color{cyan}40.5} & 37.7 \\
30 FPS- MB        & 34.0 & {\color{Purple}35.1}  & 33.2   & 23.7 & 28.0 & 29.2 & 21.7 & 14.2  & 29.5 & {\color{green}40.3} & 29.6 & {\color{red}42.9}  & {\color{blue}40.1}      & {\color{cyan}39.7} & 33.4\\
240 FPS           & {\color{red}49.5} & {\color{Purple}47.1}  & 45.3   & 34.3 & 44.8 & 43.9 & 33.3 & 19.9  & 39.5 & {\color{green}47.8} & 37.5 & {\color{cyan}47.3}  & {\color{blue}47.7}      & 46.9 & 38.6\\
\hline Speed (CPU) & {\color{cyan}38.3} & 3.8 & {\color{blue}50.8} & 10.0 & 12.5 & {\color{Purple}16.6} & {\color{red}170.4} & {\color{green}85.1} & 10.8 & 9.7 & 11.1 & 0.7 & 2.5 & 3.2 & 3.9\\
Speed (GPU) & - & - & - & - & - & - & - & - & - & {\color{cyan}43.1} & - & {\color{Purple}2.6} & {\color{blue}48.2} & {\color{green}51.8} & {\color{red}155.3}\\
\hline
\end{tabular}
}
\end{table*}

\subsection{Per Tracker Evaluation}

Figure~\ref{fig:OP_each_tracker} compares tracking higher versus lower frame rate videos for each evaluated method. For lower frame rate tracking (30 FPS) results are reported for both with and without motion blur. All CF trackers achieve a significant increase in performance (AUCs are improved $>$ 10$\%$) when tracking on 240 FPS videos. This is because in higher frame rate video, the appearance change between two adjacent frames is very small, which can be effectively learned by per frame CF online adaptation. Among deep trackers, FCNT achieved the most improvement (6$\%$), since this tracker also fine-tunes every 20 frames using the most confident tracking results. The lowest improvement belongs to SFC and MDNet. These methods are trained off-line and do not update a model or maintain a memory of past appearances~\cite{bertinetto2016fully}. Thus, tracking higher frame rate videos offers much smaller improvement to such trackers. When evaluating lower frame rate videos, a slight performance drop can be observed with the presence of motion blur, demonstrating that all trackers are reasonably robust towards motion blur.

\subsection{Overall Comparison}

The overall comparison of all trackers over three tracking settings- higher frame rate tracking (240 FPS), lower frame rate tracking with synthesized motion blur (30 FPS MB) and lower frame rate tracking without motion blur (30 FPS no MB)- is demonstrated in Fig.~\ref{fig:OP_sep} (success plots) and Table~\ref{table:all_trackers_AUC} (AUCs and tracking speed). 

\qsection{Accuracy Comparison} For lower frame rate tracking without motion blur MDNet achieved the best performance followed by SFC. HDT which utilizes deep features over a CF framework obtained the third rank followed by FCNT. Almost the same ranking is observed for lower frame rate tracking with motion blur. 
Overall, deep trackers outperformed CF trackers for lower frame rate tracking. This is not surprising, as deep trackers have a high learning capacity and employ highly discriminative deep features which are able to handle the large variation in adjacent frames which is present in lower frame rate videos.

Surprisingly, the best accuracy of higher frame rate tracking achieved by BACF (49.56), which is a CF tracker with HOG features, followed by HDT (47.80)- a CF tracker with deep features. SRDCF (47.13), Staple (45.34), DSST (44.80) and SAMF (43.92) outperformed GOTURN (38.65) and obtained very competitive accuracy compared to other deep trackers including SFC (47.78), MDNet (47.34) and FCNT (46.94). This implies that when higher frame rate videos are vailable, the ability of CF trackers to adapt online is of greater benefit than high learning capacity of deep trackers. The reasoning for this is intuitive, since for higher frame rate video there is less appearance change among consecutive frames, which can be efficiently modeled by updating the tracking model at each frame even using simple hand-crafted features.

\qsection{Run-time Comparison} The tracking speed of all evaluated methods in FPS is reported in Table~\ref{table:all_trackers_AUC}. For the sake of fair comparison, we tested MATLAB implementations of all methods (including deep trackers) on a 2.7 GHz Intel Core i7 CPU with 16 GB RAM. We also reported the speed of deep trackers on nVidia GeForce GTX Titan X GPU to have a better sense of their run-time when GPUs are available. On CPUs, all CF trackers achieved much higher speed compared to all deep trackers, because of their shallow architecture and efficient computation in the Fourier domain. %Deep trackers on CPUs showed much slower tracking speed which is only comparable with SRDCF (3.8 FPS) among all CF trackers
Deep trackers on CPUs performed much slower than CFs, with the exception of SRDCF (3.8 FPS). 
On GPU, however, deep trackers including GOTURN (155.3 FPS), FCNT (51.8 FPS) and SFC (48.2 FPS) %achieved much faster tracking speed than CPUs, which is comparable with the CPU speed of CF trackers such as KCF (170.4 FPS), BACF (38.3 FPS) and Staple (50.8 FPS). For tracking lower frame rate videos, only BACF, Staple, KCF and CFLB can perform real time on CPUs. Moreover, GOTURN, FCNT and SFC are deep trackers which offer real-time tracking of lower frame rate videos on GPUs. KCF (170.4 FPS) and GOTURN (155.3 FPS) are the only trackers which can track higher frame rate videos almost real-time on CPUs and GPUs, respectively.
performed much faster than on CPU. Their performance is comparable with many CF trackers running on CPU, such as  KCF (170.4 FPS), BACF (38.3 FPS) and Staple (50.8 FPS).
For tracking lower frame rate videos, only BACF, Staple, KCF and CFLB can track at or above real time on CPUs. GOTURN, FCNT and SFC offer real-time tracking of lower frame rate videos on GPUs. KCF (170.4 FPS) and GOTURN (155.3 FPS) are the only trackers which can track higher frame rate videos almost real time on CPUs and GPUs, respectively.

\begin{figure}
\begin{center}
    \begin{tabular}{c}
    {\small{SRDCF (soccer ball)}} \\
    \includegraphics[width=.465\textwidth]{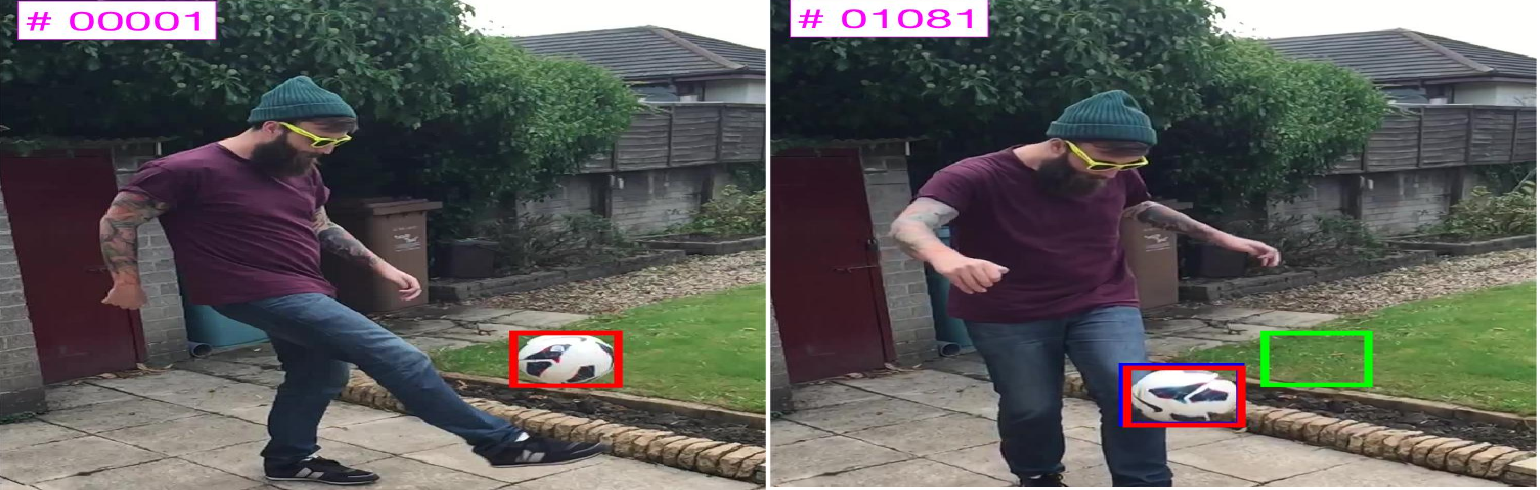} \\
   
    {\small{HDT (pingpong)}}  \\
    \includegraphics[width=.465\textwidth]{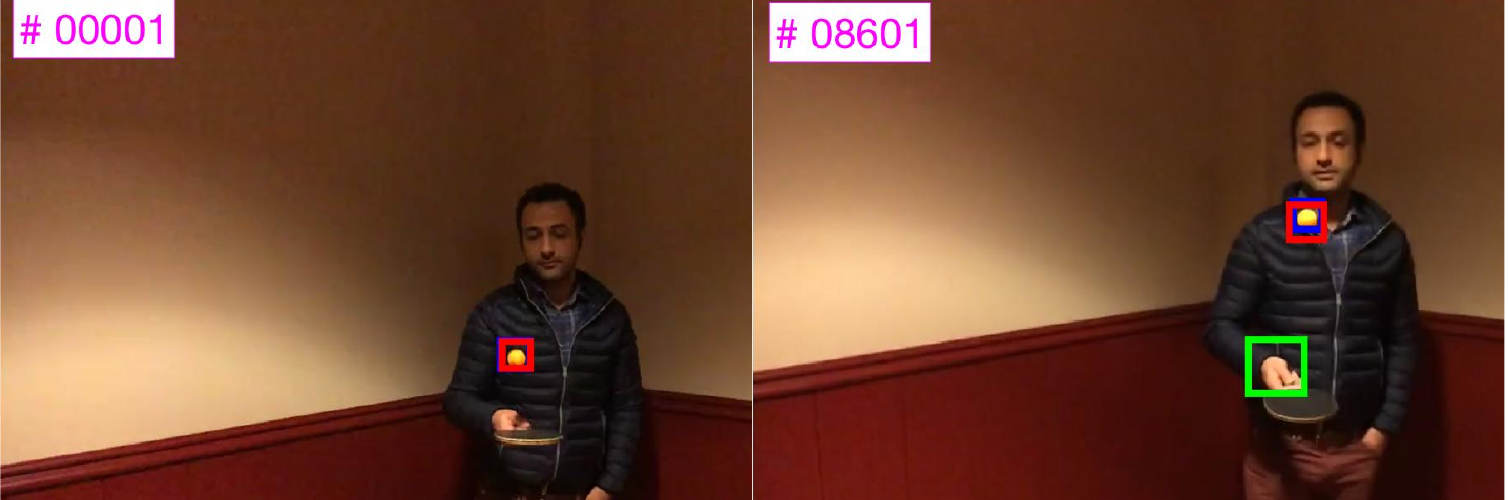} \\
    
    {\small{MDNet (tiger)}} \\
    \includegraphics[width=.465\textwidth]{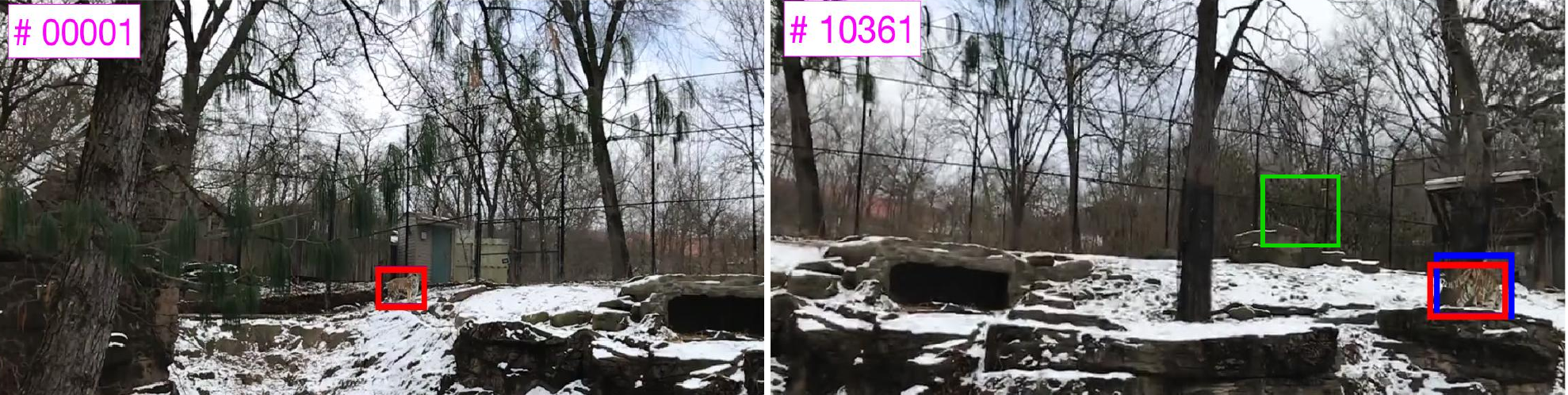} \\

    {\small{failure case (fish)}} \\
    \includegraphics[width=.465\textwidth]{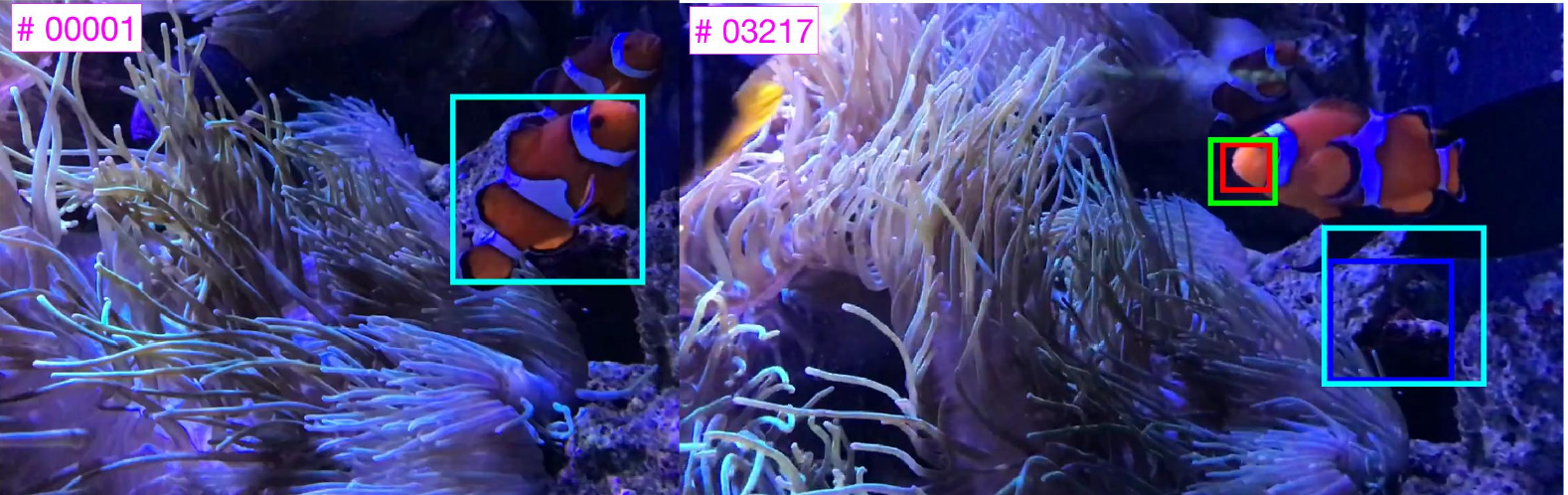} \\

    \end{tabular}
\end{center}
            \caption{Rows(1-3) show tracking performance of three trackers icluding a CF tracker with HOG (SRDCF), a CF tracker with deep features (HDT) and a deep tracker (MDNets), comparing lower frame rate ({\color{green}green} boxes) versus higher frame rate ({\color{red}red} boxes) tracking. Ground truth is shown by {\color{blue}blue} boxes. Last row visualizes a failure case of higher frame rate tracking caused by non-rigid deformation for {\color{red}BACF}, {\color{green}Staple}, {\color{blue}MDNet} and {\color{cyan}SFC}.}
\label{fig:tracking_results}
\end{figure}

\begin{figure*}
\begin{center}
    \begin{tabular}{@{}c@{} @{}c@{} @{}c@{}}
    
    \multicolumn{3}{c}{\includegraphics[width=0.4\textwidth]{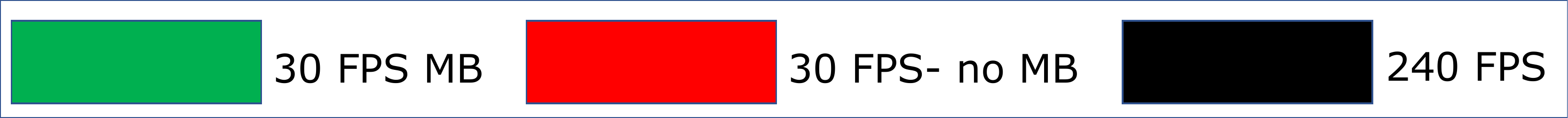}} \\
    {\small{Background clutter}} & {\small{Scale variation}} & {\small{Occlusion}}\\
    
    \includegraphics[width=0.32\textwidth]{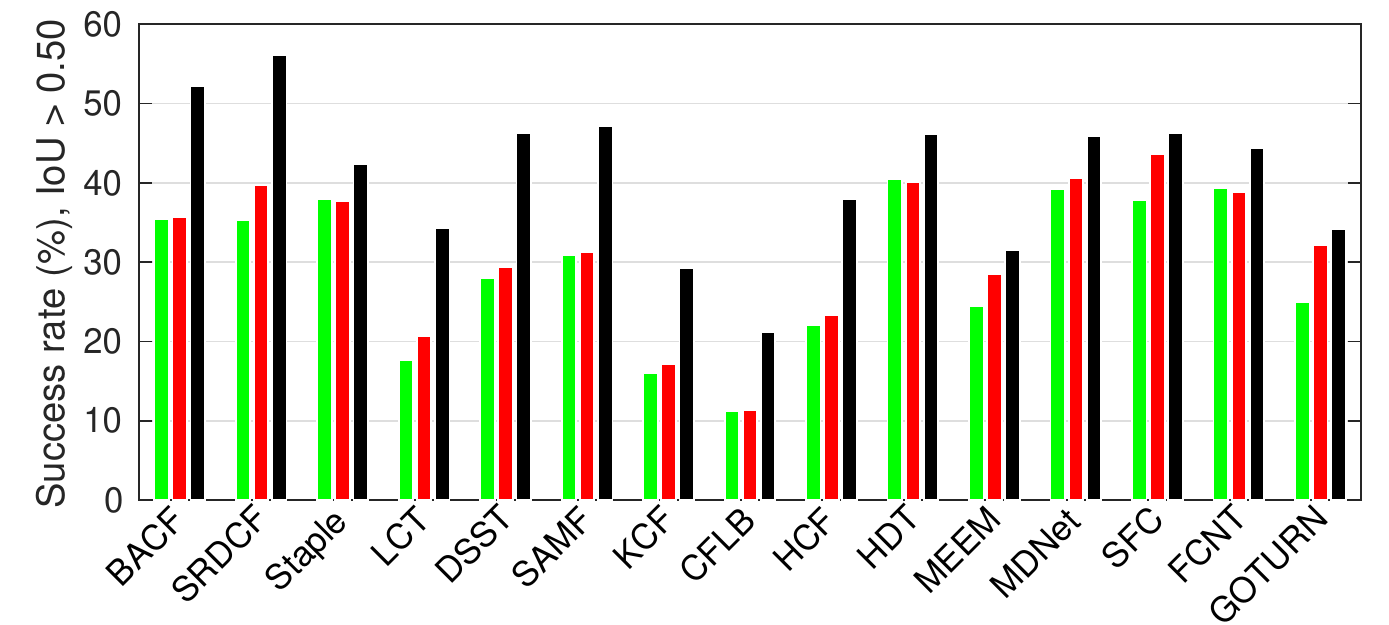} &
    \includegraphics[width=0.32\textwidth]{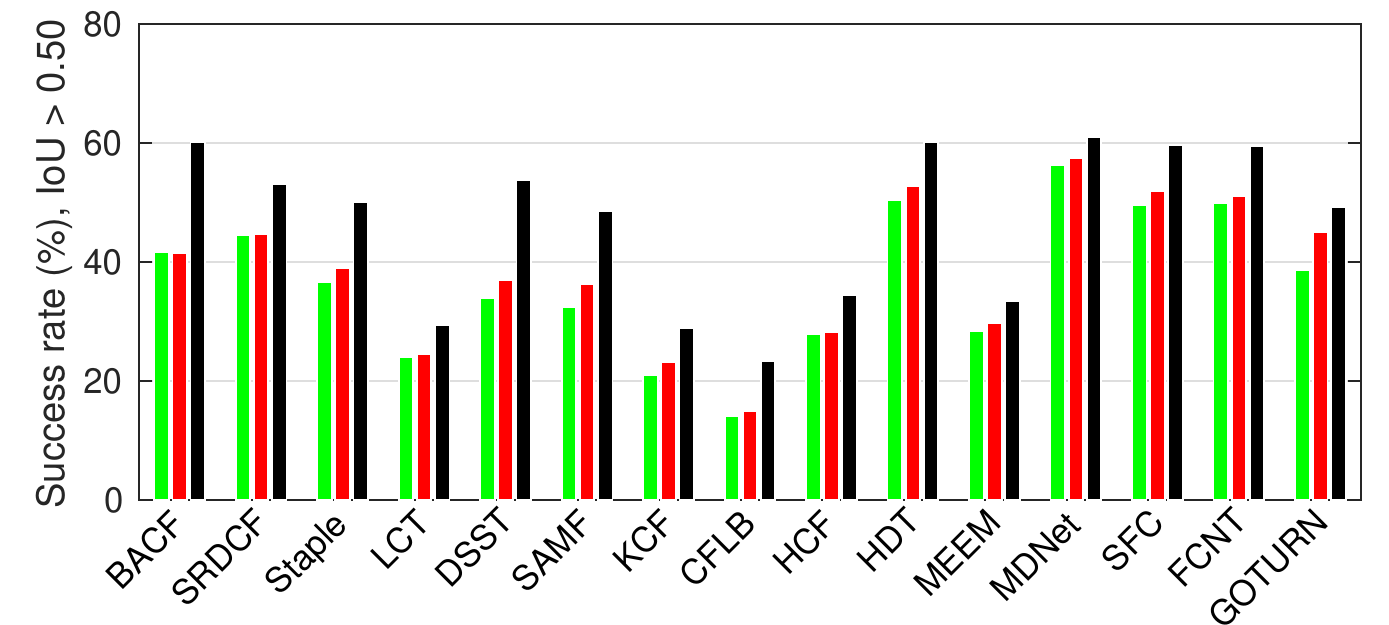} &
      \includegraphics[width=0.32\textwidth]{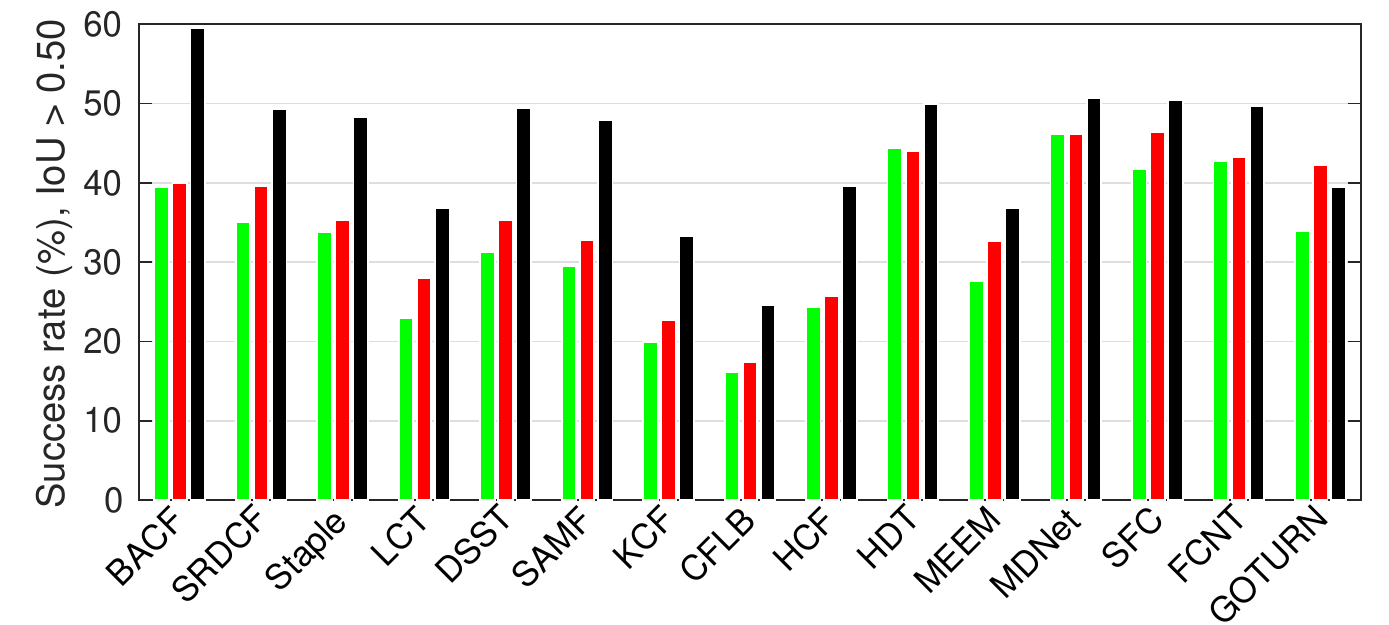}\\

      {\small{Illumination variation}} & {\small{Fast motion}} & {\small{Out-of-view}}\\
        \includegraphics[width=0.32\textwidth]{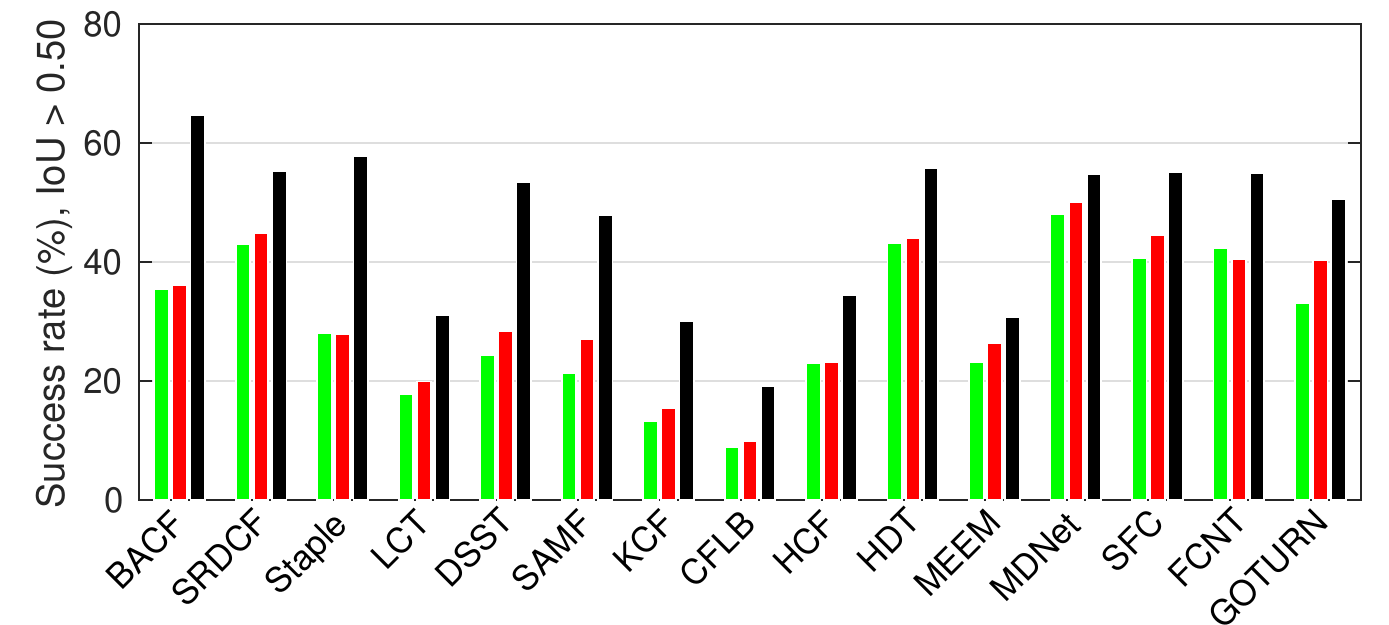} &
    \includegraphics[width=0.32\textwidth]{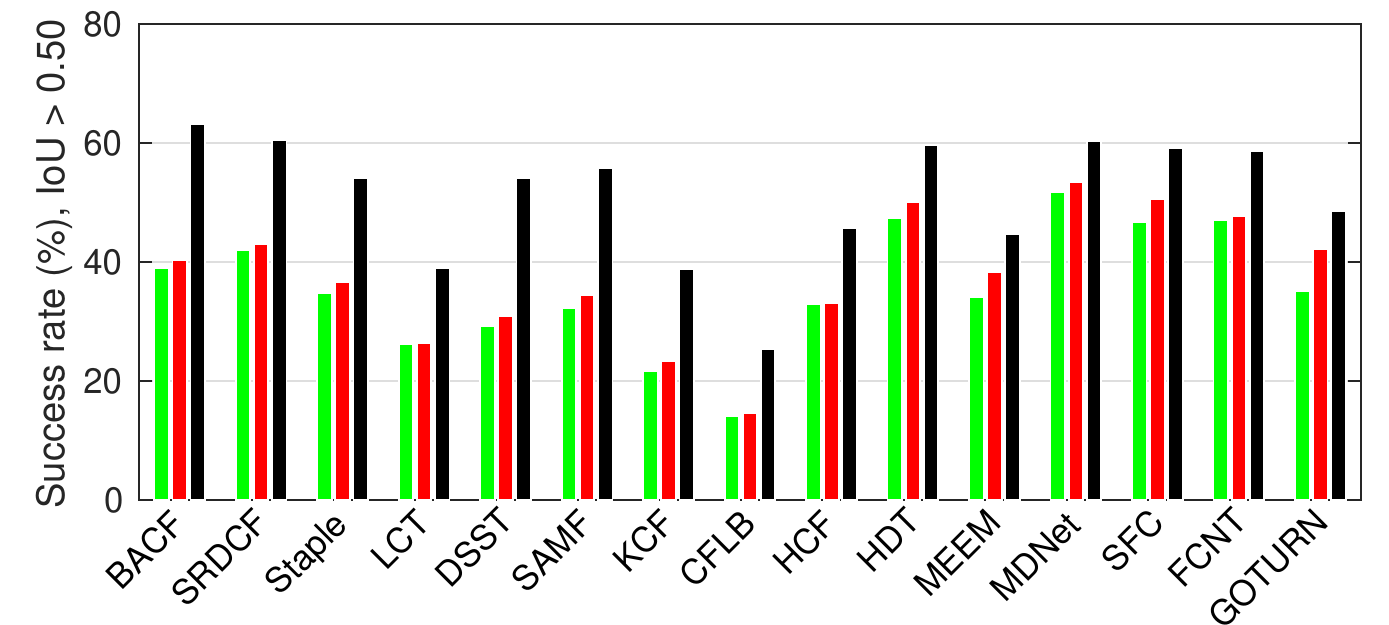} &
      \includegraphics[width=0.32\textwidth]{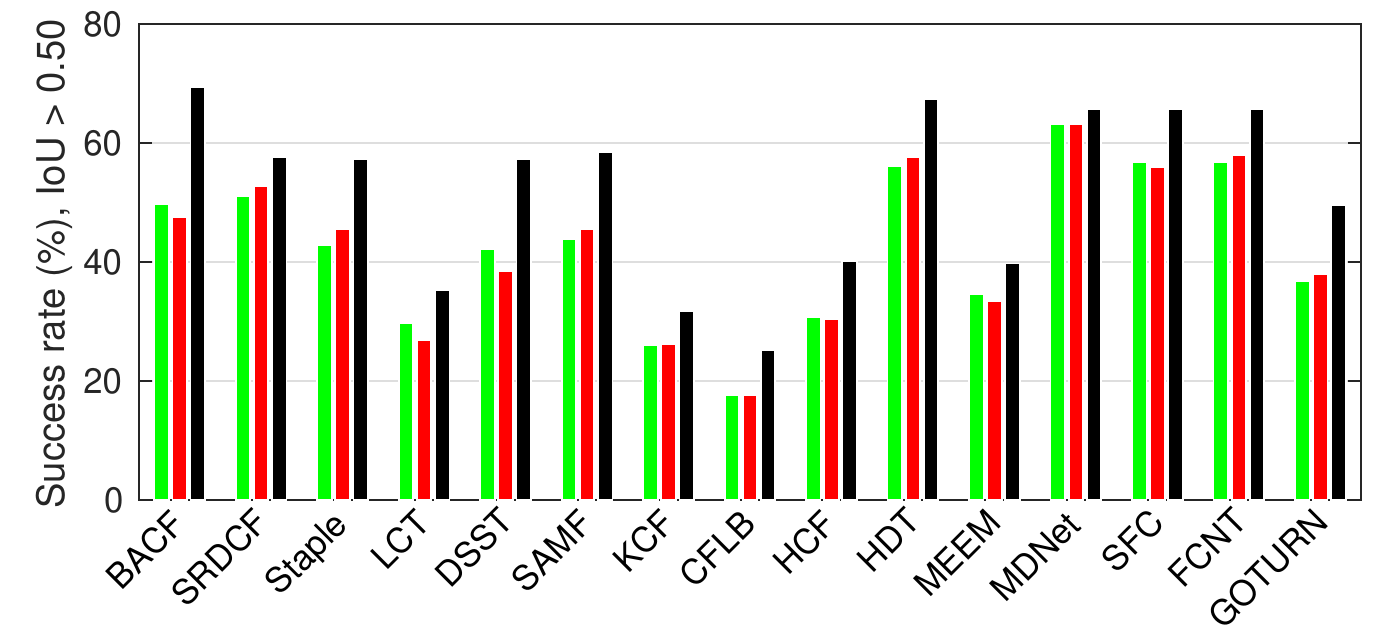}\\

      {\small{Deformation}} & {\small{Viewpoint change}} & {\small{Low resolution}}\\
        \includegraphics[width=0.32\textwidth]{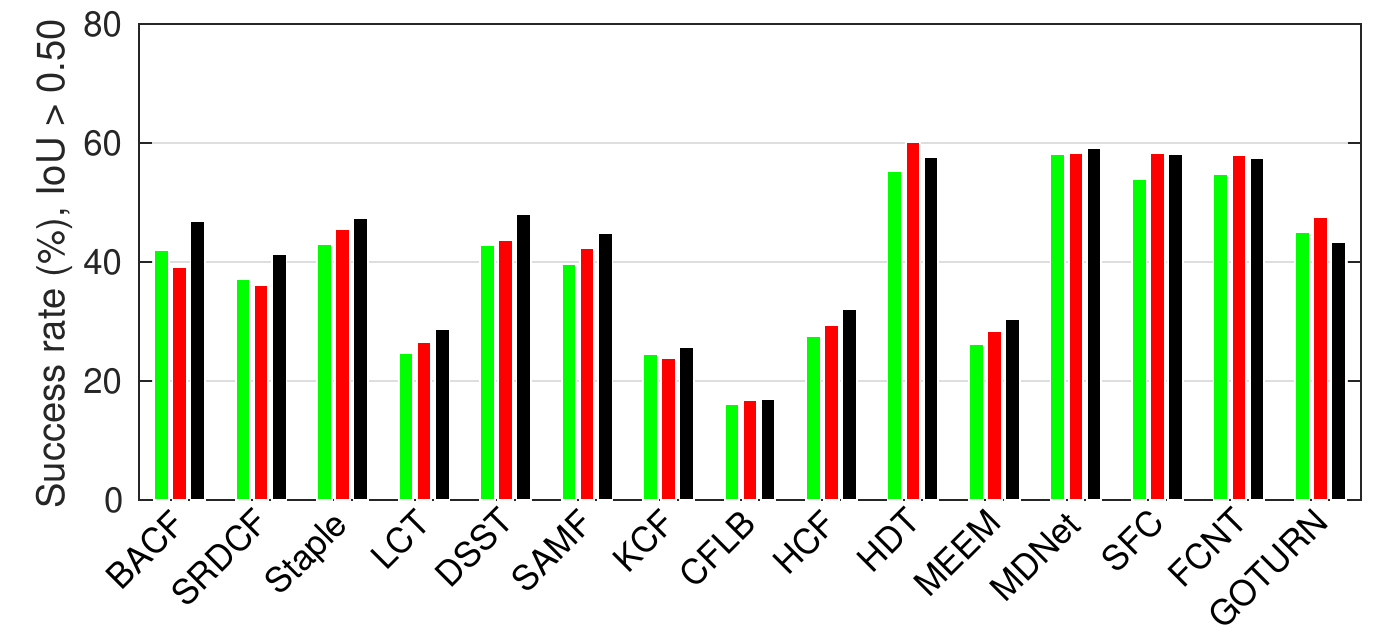} &
    \includegraphics[width=0.32\textwidth]{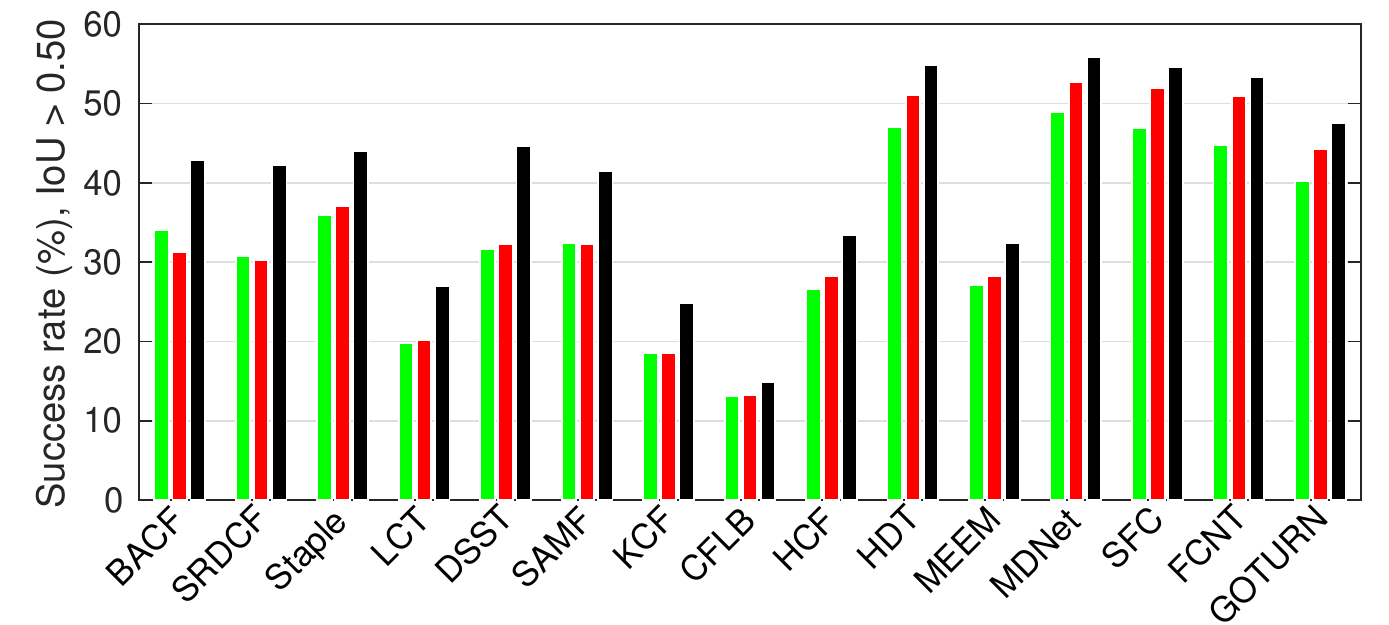} &
      \includegraphics[width=0.32\textwidth]{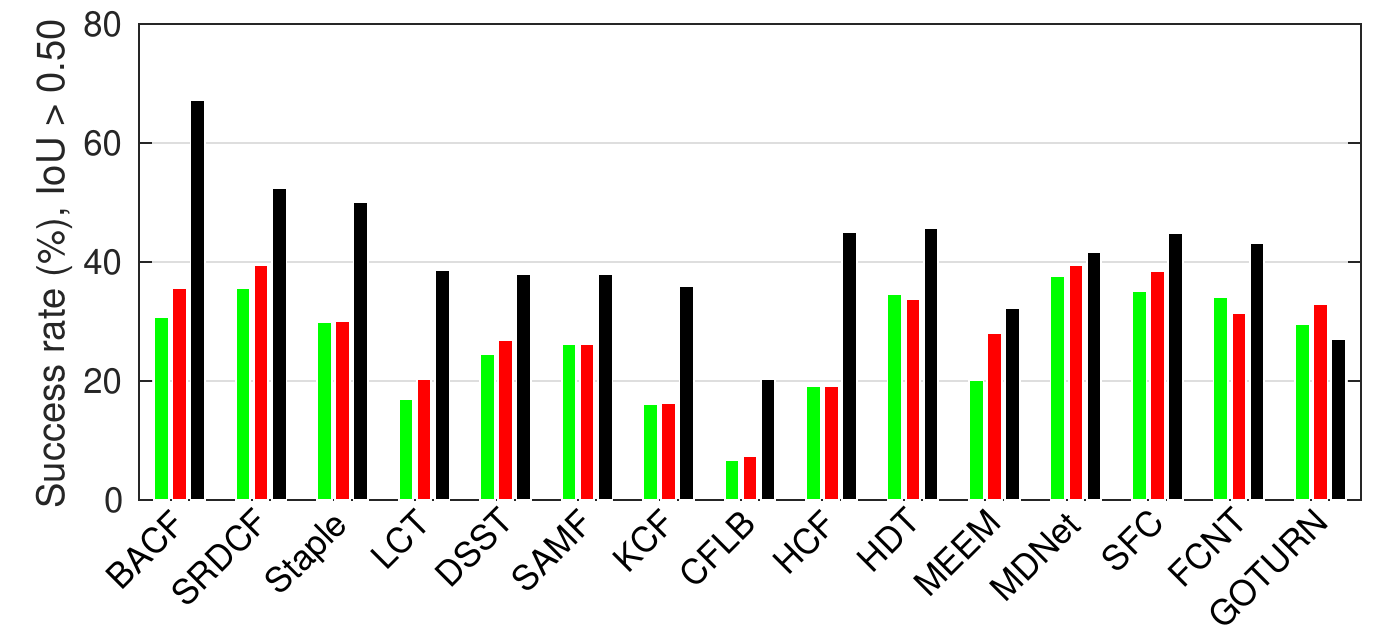}\\

    \end{tabular}
\end{center}
            \caption{Attribute based evaluation. Results are reported as success rate ($\%$) at IoU $>$ 0.50. }
\label{fig:OP_attr}
\end{figure*}

\begin{table}
\centering
\caption{Relative accuracy improvement ($\%$) of high frame rate tracking versus low frame rate tracking for each attribute. Relative improvement more than 50$\%$ are underlined in red.}
\label{table:reative_imp_attr}
\resizebox{\linewidth}{!}{
\begin{tabular}{l@{\hskip .4em }l@{\hskip .4em }l@{\hskip .4em }l@{\hskip .4em }l@{\hskip .4em }l@{\hskip .4em }l@{\hskip .4em }l@{\hskip .4em }l@{\hskip .4em }l@{\hskip .4em }}
\hline
~ & IV & SV & OCC & DEF & FM & VC & OV & BC & LR \\ \hline
  BACF & \myul[red]{82.1}  & 44.5  & \myul[red]{50.8} &  11.7     & \myul[red]{62.1}  & 26.1  & 39.6  & 47.2 & \myul[red]{118.3} \\
  SRDCF & 28.4 &  19.1  & 40.7  & 11.3    &  43.9  & 36.9 &  13.0 &  \myul[red]{59.3} & 47.2\\
  Staple & \myul[red]{106.0}  & 37.1 &  42.7  & 10.3   &  \myul[red]{55.9}  & 22.5 &  33.8  & 11.6 & \myul[red]{66.9}\\
  LCT &  \myul[red]{74.5}  & 22.4 &  \myul[red]{60.7}  & 16.3      &  48.7  & 36.2 &  18.3  & \myul[red]{94.8}  & \myul[red]{127.8}\\
 DSST &  \myul[red]{119.4}  & \myul[red]{58.4} &  \myul[red]{57.9}  & 12.1    & \myul[red]{84.5}  & 41.1 &  36.0  & \myul[red]{65.3}    & \myul[red]{54.4}\\
 SAMF &  \myul[red]{124.4}  & 49.9 &  \myul[red]{62.6}  & 13.2    & \myul[red]{73.0}  & 28.2 &  33.1  & \myul[red]{52.6}     & 44.9\\
  KCF & \myul[red]{128.0}  & 37.7 &  \myul[red]{66.7}  &  5.1     & \myul[red]{78.7}  & 34.3 &  21.4  & \myul[red]{81.9}     & \myul[red]{123.6}\\
  CFLB &  \myul[red]{113.2}  & \myul[red]{66.7} &  \myul[red]{52.3}  &  5.4  & \myul[red]{79.1}  & 14.0 &  42.3  & \myul[red]{89.3}    & \myul[red]{200.4}\\
  HCF &  \myul[red]{50.1}  & 23.1 &  \myul[red]{62.8}  & 16.2    & 39.0  & 26.0 &  30.6  & \myul[red]{71.8}     & \myul[red]{134.5}\\ 
  HDT &  29.2  & 19.1 &  12.5  &  4.3    & 26.2  & 16.6 &  19.8  & 13.8     & 32.5\\
  MEEM &  32.2  & 17.8 &  33.6  & 16.3   & 31.1  & 19.1 &  14.9  & 29.0     & \myul[red]{60.9}\\
  MDNet &  14.1  &  8.3 &   9.7  &  1.9    & 16.5  & 13.9 &   4.1  & 16.9   & 10.6 \\ 
  SFC &  35.4  & 20.1 &  20.6  &  7.9    & 26.7  & 16.6 &  15.7  & 22.1     & 27.8\\
  FCNT & 29.7  & 19.3 &  16.1  &  4.8   & 24.8  & 19.1 &  15.5  & 12.5      &  26.5\\
  GOTURN & \myul[red]{52.9} &  27.7 &  16.5  & -3.8 &  38.3 &  18.2 &  34.9  & 37.2 & -8.6\\
  \hline

\end{tabular}
}
\end{table}

\subsection{Attribute-based Evaluation}
The attribute based evaluation of all trackers on three tracking settings is shown in Fig.~\ref{fig:OP_attr} (success rate at IoU $>$ 0.50) and Table~\ref{table:reative_imp_attr} (relative accuracy improvement). Similar to the previous evaluation, the presence of motion blur slightly degrades the performance of tracking lower frame rate sequences. For lower frame rate tracking, in general, deep trackers outperformed CF trackers over all nine attributes, MDNet outperformed SFC and FCNT, and HDT achieved the superior performance compared to other CF trackers for all attributes. Compared to lower frame rate tracking, tracking higher frame rate videos offers a notable improvement of all trackers on all attributes except non-rigid deformation, as can be seen in Table~\ref{table:reative_imp_attr}. This demonstrates the sensitivity of both CF based and deep trackers to non-rigid deformation even when they track higher frame rate videos. Fig.~\ref{fig:OP_attr} shows that CF trackers with hand-crafted features outperformed all deep trackers as well as HDT for 6 attributes. More particularly, for illumination variation, occlusion, fast motion, out-of-view, background clutter and low resolution CF trackers with hand-crafted features (such as BACF and SRDCF) achieved superior performance to all deep trackers and HDT. However, deep tracker MDNet achieved the highest accuracy for scale variation (61.0), deformation (59.2) and view change (55.9), closely followed by BACF for scale variation (60.1) and HDT for deformation (57.6) and view change (54.8). The relative accuracy improvement of tracking higher frame rate versus lower frame rate videos (with motion blur) for each tracker and each attribute is reported in Table~\ref{table:reative_imp_attr}. The result shows that that first, compared to other attributes, less improvement achieved for non-rigid deformation attribute for all trackers, and second, the percentage of relative improvement for CF trackers is much higher than that of deep trackers.

\begin{figure*}
\begin{center}

\begin{tabular}{cccc}
   \includegraphics[scale=.60]{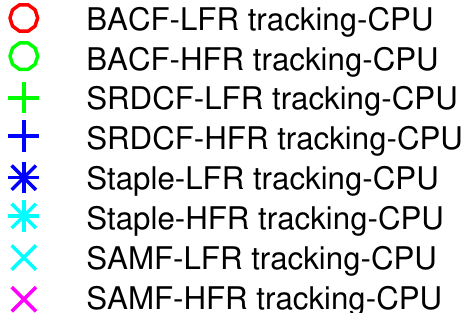}  &   
    \includegraphics[scale=.60]{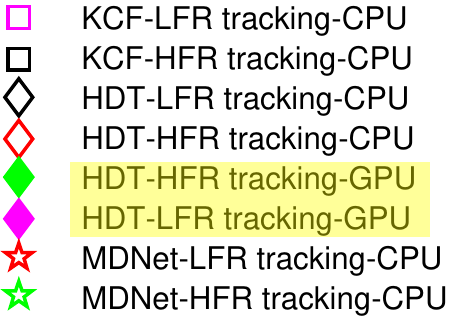} &    
     \includegraphics[scale=.60]{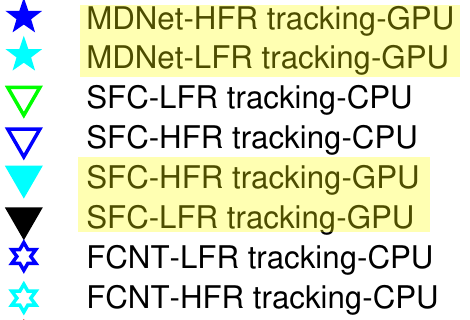} & 
      \includegraphics[scale=.60]{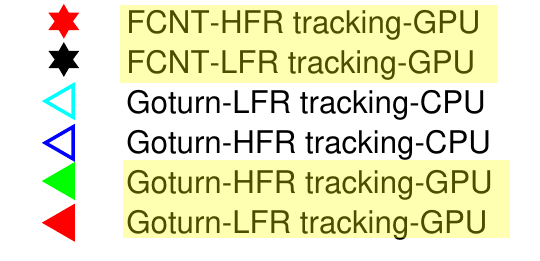} \\
\multicolumn{4}{c}{ \includegraphics[width=.8\textwidth]{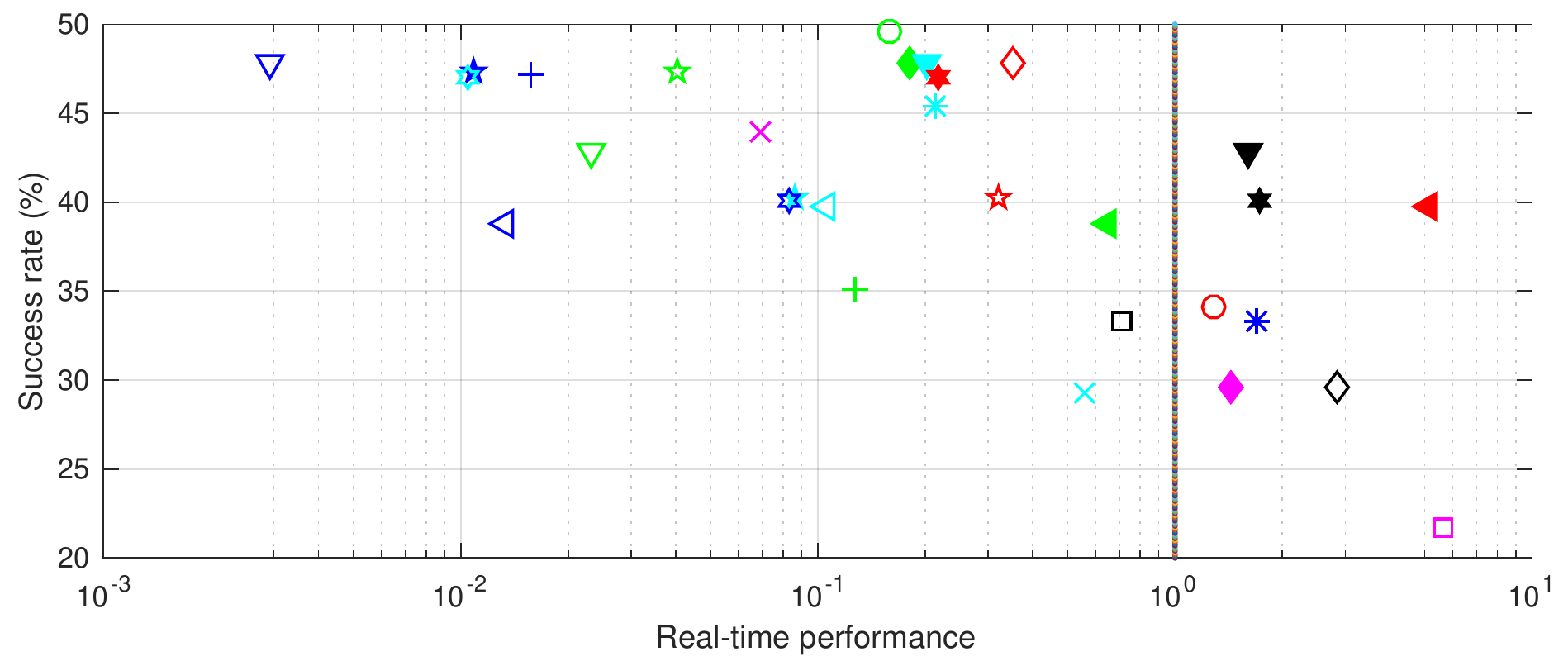}}
\end{tabular}

\end{center}
            \caption{This plot shows the affect of resource availability (GPUs vs. CPUs) and frame rate of captured videos (lower vs. higher frame rate) on the top-10 evaluated trackers' accuracies and real-time performance. Real-time performance is computed as the ratio of each tracker's speed (FPS) to frame rate of the target videos (30 vs. 240 FPS). The vertical line on the plot shows the boundary of being real-time (frame rate of the target video is the same as the tracker's speed). Trackers which are plotted at the left side of the line are not able to track real-time (according to their tracking speed and video frame rate). GPU results are highlighted in yellow.}%, those on the right side of the line perform real-time. The legend shows each tracker, tracking case (high and low frame rate tracking), and processing resource availability (GPUs and CPUs).}
\label{fig:final_diss}
\end{figure*}

\subsection{Discussion and Conclusion}
In this paper, we introduce the first higher frame rate object tracking dataset and benchmark. We empirically evaluate the performance of the state-of-the-art trackers with respect to two different capture frame rates (30 FPS vs. 240 FPS), and find the surprising result that at higher frame rates, simple trackers such as correlation filters trained on hand-crafted features (\eg HOG) outperform complex trackers based on deep architecture. This suggests that computationally tractable methods such as cheap CF trackers in conjunction with higher capture frame rate videos can be utilized to effectively perform object tracking on devices with limited processing resources such as smart phones, tablets, drones, etc. As shown in Fig.~\ref{fig:final_diss}, cheaper trackers on higher frame rate video (\eg KCF and Staple) have been demonstrated as competitive with many deep trackers on lower frame rate videos (such as HDT and FCNT).

Our results also suggest that traditional evaluation criteria that trades off accuracy versus speed (e.g., Fig.7 in ~\cite{kristan2015visual}) could possibly paint an incomplete picture. This is because, up until now, accuracy has been measured without regard to the frame rate of the video. As we show, this dramatically underestimates the performance of high speed algorithms. In simple terms: the accuracy of a 240 FPS tracker cannot be truly appreciated until it is run on a 240 FPS video! From an embedded-vision perspective, we argue that the acquisition frame rate is a resource that should be explicitly traded off when designing systems, just as is hardware (GPU vs CPU). Our new dataset allows for, the first time, exploration of such novel perspectives.
Our dataset fills a need. The need for speed.

{\small
\bibliographystyle{ieee}
\bibliography{egbib}
}

\end{document}